\documentclass[conference]{IEEEtran}
\IEEEoverridecommandlockouts
\usepackage{cite}
\usepackage{amsmath,amssymb,amsfonts}
\usepackage{graphicx}
\usepackage{textcomp}
\usepackage{xcolor}
\def\BibTeX{{\rm B\kern-.05em{\sc i\kern-.025em b}\kern-.08em
    T\kern-.1667em\lower.7ex\hbox{E}\kern-.125emX}}
\begin{document}

\title{
ChunkVLA-AM: Parallel Action Chunking for Vision-Language-Action Robot Control in Additive Manufacturing
}

\author{
{\small Zhugang Liu$^{1}$}\\
{\scriptsize\ttfamily zhugang.liu01@utrgv.edu}
\and

{\small Kaichuang Zhang$^{3}$}\\
{\scriptsize\ttfamily kaichuangzhang@usf.edu}
\and

{\small Jinman Zhang$^{1}$}\\
{\scriptsize\ttfamily jinman.zhang01@utrgv.edu}
\and

{\small Pu Sun$^{4}$}\\
{\scriptsize\ttfamily psun9932@sdsu.edu}
\and

{\small Martha Asare$^{1}$}\\
{\scriptsize\ttfamily martha.asare01@utrgv.edu}
\and

{\small Jose Hernandez$^{2}$}\\
{\scriptsize\ttfamily jose.hernandez112@utrgv.edu}
\and

{\small Maxim Ermolinsky$^{2}$}\\
{\scriptsize\ttfamily maxim.ermolinsky01@utrgv.edu}
\and

{\small Efren Saenz$^{1}$}\\
{\scriptsize\ttfamily efren.saenz01@utrgv.edu}
\and

{\small Qi Lu$^{1}$}\\
{\scriptsize\ttfamily qi.lu@utrgv.edu}
\and

{\small Jinghao Yang$^{2,*}$}\\
{\scriptsize\ttfamily jinghao.yang@utrgv.edu}
\and[\hfill\break\mbox{}\hfill]

{\footnotesize $^{1}$Department of Computer Science, The University of Texas Rio Grande Valley}\\
{\footnotesize $^{2}$Department of Electrical and Computer Engineering, The University of Texas Rio Grande Valley}\\
{\footnotesize $^{3}$Department of Electrical Engineering, University of South Florida}\\
{\footnotesize $^{4}$Department of Computer Science, San Diego State University}\\
{\footnotesize $^{*}$Corresponding Author}
}
\maketitle

\begin{abstract}
Vision–language–action (VLA) models unify visual perception, language understanding, and action generation, offering new opportunities for automation in additive manufacturing (AM). 
However, deployment in AM remains challenging because adapting these models to unseen robot embodiments is costly, and performance can degrade under environment changes.
In this work, we present a framework for deploying OpenVLA-OFT on a FAIRINO FR3 robot in a fixed AM workcell.
A data pipeline converts monocular real-world demonstrations into OpenVLA-compatible TFDS/RLDS datasets to support adaptation to the FR3 embodiment. At runtime, each inference request predicts an eight-step chunk of 7-D actions. The FR3 executes each chunk open loop before capturing a new observation, providing closed-loop feedback between chunks. The system uses a cloud–edge architecture in which the FR3 client streams observations to a remote inference server through a FastAPI interface. 
In 42 physical A-to-B object-transfer trials, evenly split between red and blue targets, the system succeeded in 39 (92.9\%). All three failures occurred during final placement, when insufficient release-height control caused the object to topple. An illumination sweep identified a low-error luminance range of 85–125 on a 0–255 scale, with the lowest mean spatial error at 95.
\end{abstract}.

\begin{IEEEkeywords}
vision-language-action model, additive manufacturing, robot manipulation, action chunking, OpenVLA
\end{IEEEkeywords}

\section{Introduction}

Additive manufacturing (AM) has become an important production paradigm for rapid prototyping, customized medical devices, aerospace components, and high-value industrial parts~\cite{Gibson2021AM,Ngo2018AMReview,DebRoy2018MetalAM}. Beyond deposition itself, practical AM workflows require inspection, part retrieval, build-plate handling, post-processing, and occasional intervention after process failures. These operations can expose human operators to process-specific risks: fused deposition modeling (FDM) may emit ultrafine particles and volatile organic compounds during heated thermoplastic extrusion~\cite{Stephens2013UFP,Azimi2016FDMEmissions}, while selective laser melting (SLM) involves fine metallic powders and particulate-matter hazards~\cite{Chen2020MetalAMHazards}. Robotic automation is therefore attractive for AM workcells because it can improve repeatability and reduce human exposure during inspection, transfer, and failure response. However, conventional rule-based robot programs are difficult to reuse across different printers, robot platforms, camera configurations, lighting conditions, object poses, and defect types such as FDM warping, inter-layer delamination, SLM porosity, lack-of-fusion holes, and cracks~\cite{Zhang2006FDMFEM,Jin2020InterlayerAI,Zhang2017SLMDefects}.

Vision-language-action (VLA) models provide a possible route toward more flexible AM robotic systems. By mapping visual observations and natural-language task descriptions directly to robot actions, VLA policies can represent tasks such as ``retrieve the printed part,'' ``inspect the warped edge,'' or ``move the build-plate sample to the cooling area'' without requiring a separate perception pipeline and a handcrafted state machine. Recent systems including RT-1, RT-2, Open X-Embodiment, and OpenVLA have shown that large-scale robot datasets, vision-language backbones, and action-token prediction can support general manipulation and transfer across tasks and embodiments~\cite{Brohan2022RT1,Brohan2023RT2,OpenXEmbodiment2023,Kim2024OpenVLA}. This capability is relevant to AM because the robot must interpret both the visual manufacturing context and the language instruction while generating manipulation commands within a constrained workspace.

Despite this potential, transferring a VLA model to an AM workcell is not a straightforward model-deployment problem. The first difficulty is migrating demonstrations and action interfaces across robot platforms. Robot datasets often differ in image resolution, observation structure, sampling rate, episode organization, coordinate frames, state definitions, and gripper conventions. Their action spaces may also represent joint positions, joint velocities, absolute Cartesian poses, or relative end-effector motion. Even when two platforms expose action vectors of the same dimension, the individual elements may have different physical meanings, units, ranges, or normalization rules. RLDS and TFDS provide a common structure for organizing episodic robot data, but they do not by themselves resolve these embodiment-specific differences~\cite{Guadarrama2021RLDS,TFDS}. As a result, transferring an existing VLA pipeline to a platform such as the FAIRINO FR3 requires more than renaming dataset fields. Coordinate transformations, action semantics, gripper commands, and normalization statistics must all be defined consistently before the model can be trained and safely deployed.


The second difficulty is robustness to changes between the demonstration environment and the deployed AM workcell. Lighting intensity, illumination direction, camera exposure, background clutter, camera viewpoint, and object pose may vary between data collection and execution. Reflective build plates, dark polymer parts, shadows cast by the robot, and changes in printer illumination can further alter the visual appearance of the part and the gripper. These variations are especially relevant when the model is adapted using demonstrations collected under a fixed camera and lighting configuration. Because a VLA policy predicts actions directly from image observations, it may associate the required motion with incidental visual features of the training environment instead of the task geometry. A policy that succeeds under the original lighting condition may therefore become less reliable when the workcell appearance changes, even though the physical task remains the same.


In this work, we present \emph{ChunkVLA-AM}, a reproducible OpenVLA-OFT-based deployment pipeline for FR3-based AM post-print retrieval. To make embodiment adaptation explicit, we construct a conversion pipeline that transforms monocular FR3 demonstrations, Cartesian end-effector states, binary gripper commands, language instructions, and episode metadata into OpenVLA-compatible TFDS/RLDS episodes. The pipeline defines the coordinate frames, action semantics, and normalization procedure required by the target FR3 system, while LoRA is used to avoid full-parameter fine-tuning. For robot control, we adopt a horizon-$H$ action-chunk interface and use $H=8$ in the reported physical tests, allowing the model to predict short-horizon motion sequences in a single request. The FR3 client then applies action denormalization, workspace clipping, unsafe-action rejection, and chunk-wise replanning after execution of the returned actions. This implementation focuses on the system-level adaptation and deployment required to use an action-chunked VLA policy in a fixed AM workcell and to evaluate its behavior under changes in lighting and object pose.
\section{Related Work}
\label{sec:related}

\subsection{Vision-Language-Action Models for Robot Control}

Vision-language-action (VLA) models have emerged from the broader trend of connecting language, perception, and embodied decision making. Early generalist and language-conditioned robot systems, including Gato, SayCan, PaLM-E, and VIMA, showed that multimodal representations can support instruction following, planning, and interactive manipulation across diverse tasks~\cite{Reed2022Gato,Ahn2022SayCan,Driess2023PaLME,Jiang2022VIMA}. More recent VLA policies such as RT-1, RT-2, Open X-Embodiment, and OpenVLA further demonstrate that large-scale robot datasets and vision-language backbones can be used to predict robot actions directly from images and natural-language instructions~\cite{Brohan2022RT1,Brohan2023RT2,OpenXEmbodiment2023,Kim2024OpenVLA}. These studies provide the main methodological foundation for applying VLA models beyond household or tabletop manipulation. However, their evaluation settings are still largely centered on general manipulation benchmarks, leaving manufacturing-oriented scenarios comparatively underexplored.

\subsection{Robotic Automation in Additive Manufacturing}

Additive manufacturing (AM) has been widely studied as a flexible production technology for customized, lightweight, and high-value components~\cite{Gibson2021AM,Ngo2018AMReview,DebRoy2018MetalAM}. Robotic systems have also been introduced into AM workflows for material handling, inspection, deposition assistance, post-processing, and large-scale or wire-arc manufacturing~\cite{Ding2015WireArcAM}. Compared with conventional automation, AM workcells pose distinctive perception and decision-making challenges because part geometry, surface appearance, thermal history, defect morphology, and process state can vary significantly across builds. Defects such as warping, delamination, porosity, lack of fusion, and cracking further require robots to reason over visual manufacturing evidence instead of simply replaying fixed trajectories~\cite{Zhang2006FDMFEM,Jin2020InterlayerAI,Zhang2017SLMDefects}. These characteristics make AM a natural application domain for VLA models: language can specify task intent, visual input can capture part and process context, and action prediction can support flexible intervention.

\subsection{Challenges of Applying VLA Models to AM}

Although VLA models are promising for AM, existing literature suggests two major barriers to practical use. The first is data and adaptation cost. AM tasks are weakly represented in large robot-learning datasets, while effective VLA adaptation requires demonstrations that align visual observations, language instructions, action labels, and episode structure in reusable formats such as TFDS/RLDS~\cite{Guadarrama2021RLDS,TFDS}. Full fine-tuning of large VLA models can be expensive and data-hungry, motivating parameter-efficient adaptation methods such as LoRA~\cite{Hu2022LoRA}. The second challenge is achieving temporally coherent motion. Single-step autoregressive prediction can accumulate covariate shift and produce open-loop drift during precise manipulation. Prior work on action chunking and horizon-based visuomotor policies shows that predicting multiple future actions as a structured sequence improves motion consistency~\cite{Zhao2023ACT,Chi2023DiffusionPolicy}. In AM settings, such temporal coherence must be balanced against responsiveness and safety, because robot motions often occur near printed parts, build plates, tools, or process-sensitive regions. Therefore, the VLA--AM research gap is not only whether VLA models can represent AM tasks, but also how they can be adapted and executed with sufficient temporal stability for precise manufacturing intervention.


\section{Preliminaries}
\label{sec:preliminaries}

\subsection{VLA Policy for AM Robot Intervention}

We formulate AM robot intervention as language-conditioned action prediction. At time $t$, the robot observes a RGB image $I_t\in\mathbb{R}^{h\times w\times 3}$ and receives a natural-language instruction $l$. A VLA policy parameterized by $\theta$ predicts a robot action from the image--language pair,
\begin{equation}
    a_t=f_{\theta}(I_t,l),
\end{equation}
where $a_t\in\mathbb{R}^{d_a}$ denotes the continuous control command after token decoding and denormalization. For an AM demonstration dataset
\begin{equation}
    \mathcal{D}_{\mathrm{AM}}=\{(I_t^{(i)},l^{(i)},a_t^{(i)},e_t^{(i)})\}_{i,t},
\end{equation}
where $e_t^{(i)}$ stores episode metadata, the adaptation objective can be written as empirical risk minimization, following standard supervised imitation-style VLA adaptation~\cite{Brohan2022RT1,Kim2024OpenVLA},
\begin{equation}
    \theta^{\star}=\arg\min_{\theta}\frac{1}{|\mathcal{D}_{\mathrm{AM}}|}
    \sum_{(I,l,a)\in\mathcal{D}_{\mathrm{AM}}}\mathcal{L}\big(f_{\theta}(I,l),a\big).
\end{equation}

\subsection{Costly Adaptation and Fine-Tuning}

Let $\theta_0$ denote a pretrained VLA model and let $N_{\theta}$ be the number of trainable parameters under full fine-tuning. Direct adaptation updates all model weights,
\begin{equation}
    \theta=\theta_0+\Delta\theta,\qquad \Delta\theta\in\mathbb{R}^{N_{\theta}},
\end{equation}
which creates a high optimization and storage cost when $N_{\theta}$ is large. The adaptation cost can be abstracted as
\begin{equation}
    C_{\mathrm{adapt}} \propto N_{\mathrm{train}}\,|\mathcal{D}_{\mathrm{AM}}|\,E,
\end{equation}
where $N_{\mathrm{train}}$ is the number of trainable parameters and $E$ is the number of training epochs. In AM, this cost is amplified by the limited availability of task-specific demonstrations and by embodiment-dependent action conventions. Parameter-efficient adaptation reduces the trainable parameter count by freezing $\theta_0$ and learning a low-dimensional update $\Delta\theta_{\phi}$~\cite{Hu2022LoRA},
\begin{equation}
    \theta=\theta_0+\Delta\theta_{\phi},\qquad |\phi|\ll N_{\theta}.
\end{equation}
Thus, the goal is to minimize task loss on $\mathcal{D}_{\mathrm{AM}}$ while keeping $N_{\mathrm{train}}=|\phi|$ small. This motivates a data interface that preserves AM observations, instructions, actions, and episode boundaries, together with parameter-efficient fine-tuning while avoiding full-model adaptation.

\subsection{Visual Input Conditions: Illumination and Camera Viewpoint}

The observed image $I_t$ depends not only on the scene content but also on physical imaging conditions. Let $\mathcal{S}_t$ denote the geometric and material state of the AM workspace at time $t$, let $\lambda_t \in \Lambda$ be the ambient illumination---characterized by mean luminance $\mu(\lambda_t)\in[0,255]$ and color temperature $T(\lambda_t)$ in Kelvin---and let $\mathbf{c}=(K,[R\mid\mathbf{t}])$ denote the camera intrinsic matrix $K$ and fixed extrinsic pose $[R\mid\mathbf{t}]$. The image formation can then be written as
\begin{equation}
    I_t = \mathcal{G}(\mathcal{S}_t,\,\lambda_t;\,\mathbf{c}),
\end{equation}
where $\mathcal{G}$ is the image formation function. In our deployment $\mathbf{c}$ is held constant across all episodes (fixed monocular camera), so cross-episode variation in $I_t$ arises primarily from changes in $\mathcal{S}_t$ (part pose, build-plate state) and $\lambda_t$ (workcell lighting).

For the VLA policy to generalize across AM workcell conditions, it must satisfy approximate illumination invariance,
\begin{equation}
    \bigl\|f_\theta\!\bigl(\mathcal{G}(\mathcal{S}_t,\lambda;\mathbf{c}),l\bigr)
    -f_\theta\!\bigl(\mathcal{G}(\mathcal{S}_t,\lambda_0;\mathbf{c}),l\bigr)\bigr\|_2
    \leq \epsilon_\lambda,\quad \forall\,\lambda\in\Lambda,
\end{equation}
where $\lambda_0$ is a reference (baseline) illumination and $\epsilon_\lambda$ is a tolerable action deviation. A small $\epsilon_\lambda$ across a wide range of $\Lambda$ indicates that the visual backbone has learned illumination-robust features.

Under a fixed camera pose $\mathbf{c}$, the appearance of the same workspace also varies with object placement and orientation on the build plate. The perspective projection of the scene can be written as
\begin{equation}
    I_t = \Pi(\mathcal{S}_t;\,K,R,\mathbf{t}),
\end{equation}
where $\Pi$ denotes the perspective projection operator. Because the camera pose is fixed, residual viewpoint variation arises from differences in printed-part pose. The policy must therefore handle object-level pose variation without explicit 3D estimation. This motivates the use of square-padded, fixed-resolution image inputs and the robustness study in Section~\ref{sec:experiments}, where we evaluate the adapted policy across seven distinct illumination conditions spanning luminance.



\subsection{Action Representation}

Each action is represented as a Cartesian end-effector delta and gripper command,
\begin{equation}
    a_t=[\Delta x,\Delta y,\Delta z,\Delta \phi,\Delta \theta,\Delta \psi,g].
\end{equation}
The first six terms describe translational and rotational increments, and $g$ is a binary gripper state.

\section{ChunkVLA-AM}
\label{sec:method}
\begin{figure*}[t]
  \centering
  \includegraphics[width=0.95\linewidth]{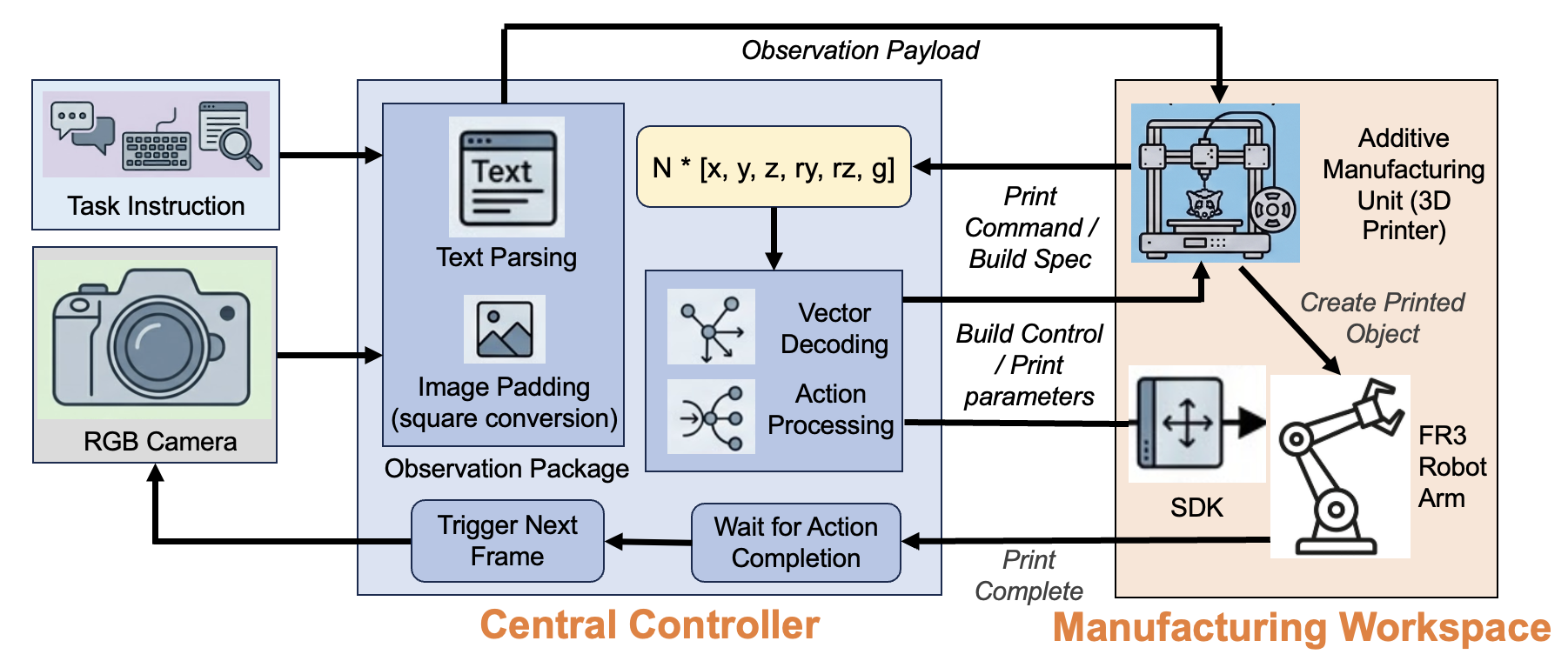}
  \caption{Cloud-edge deployment architecture. The local FR3 client sends instruction-conditioned RGB observations to a remote OpenVLA-OFT service, receives an action chunk, and validates commands before execution.}
  \label{fig:pipeline}
\end{figure*}
ChunkVLA-AM is a reproducible OpenVLA-OFT-based deployment pipeline. It uses monocular RGB input, synchronized robot poses, parallel action-chunk prediction, Cartesian end-effector control, and robot-side safety filters. The pipeline defines five interfaces:
\begin{enumerate}
    \item TFDS/RLDS conversion,
    \item Action Normalization and Token Targets,
    \item LoRA Adaptation,
    \item Action Chunk Construction
    \item Cloud-Edge Execution and Safety Filtering
\end{enumerate}

Figure~\ref{fig:pipeline} summarizes the pipeline. The remote server performs large language model inference, while the CPU-only robot workstation handles capture, requests, parsing, denormalization, smoothing, clipping, validation, execution, and logging. This split supports hardware where the robot controller and VLA model cannot run on the same machine.

\subsection{TFDS/RLDS Data Conversion} \label{subsec:4.1}

Each demonstration becomes an RLDS-style episode with synchronized observations, actions, language, and metadata~\cite{Guadarrama2021RLDS,TFDS}. The raw log contains time-stamped end-effector poses, gripper states, controller commands, and camera frames. We use the robot base frame for Cartesian translations. At step $t$, the translational action is
\begin{equation}
    \Delta p_t=p_{t+1}-p_t,
\end{equation}
where $p_t=[x_t,y_t,z_t]$ is the end-effector position. Rotation uses a roll-pitch-yaw Euler increment,
\begin{equation}
    \Delta r_t=[\Delta \phi_t,\Delta \theta_t,\Delta \psi_t],
\end{equation}
computed from consecutive orientations after conversion to the same Euler convention. The gripper command $g_t$ is binary, with $g_t=1$ for closing or holding and $g_t=0$ for opening or release. The final 7-D action is
\begin{equation}
    a_t=[\Delta x_t,\Delta y_t,\Delta z_t,\Delta \phi_t,\Delta \theta_t,\Delta \psi_t,g_t].
\end{equation}

Images are converted to RGB, square-padded, and resized to $224\times224$ pixels. Images and robot states are synchronized by nearest time stamp; frames outside one control period are dropped. The demonstrations and corresponding action labels are sampled at approximately 25~Hz. Episode boundaries come from the start/stop signal, timeout, or task-completion marker. The schema stores image observations, Cartesian state and gripper metadata, normalized 7-D delta actions, repeated language instructions, and trial metadata including episode id, step index, timestamps, and terminal flag.

\subsection{Action Normalization and Token Targets}

Before tokenization, each continuous action dimension is clipped to a symmetric range and normalized to $[-1,1]$:
\begin{equation}
    \bar{a}_{t,j}=\mathrm{clip}(a_{t,j},-c_j,c_j)/c_j.
\end{equation}
For the post-print retrieval configuration, the default clipping values are $c_x=c_y=c_z=5$~mm per action and $c_{\phi}=c_{\theta}=c_{\psi}=0.05$~rad per action. The gripper channel is not normalized as a continuous value; it is binarized by the open/close state and serialized as a discrete command token. These thresholds are deployment parameters and should not be interpreted as safety guarantees: they limit the largest command passed to the controller, but physical safety still depends on validation, contact monitoring, and emergency stop handling.

\subsection{LoRA Adaptation}

The adaptation uses only monocular RGB images and language instructions, avoiding multi-view cameras, depth sensing, tactile sensing, or other additional modalities. We adapt the OpenVLA-OFT model with LoRA, which trains low-rank updates while freezing the pretrained backbone~\cite{Hu2022LoRA}. For FR3 adaptation, we use a LoRA rank of 32 with zero dropout, a learning rate of $5\times10^{-4}$, a batch size of 8, and gradient accumulation over 4 steps, giving an effective batch size of 32. The model is trained for approximately 30 epochs.

For single-step imitation, the supervised objective is
\begin{equation}
    \mathcal{L}_{\text{step}}=-\sum_t \log p_{\theta}(a_t\mid I_t,l).
\end{equation}
For chunked imitation, adjacent expert actions form a short-horizon target,
\begin{equation}
    \mathcal{L}_{\text{chunk}}=-\sum_t\sum_{h=0}^{H-1}\log p_{\theta}(a_{t+h}\mid I_t,l).
\end{equation}
Terminal chunks are truncated to the remaining valid actions and padded positions are masked from the loss.

\subsection{Action Chunk Construction}

For a horizon $H$, the training target concatenates $H$ future normalized 7-D actions into one longer target sequence,
\begin{equation}
    C_t=[\bar{a}_t,\bar{a}_{t+1},\ldots,\bar{a}_{t+H-1}].
\end{equation}
At each physical inference cycle, the RGB image is synchronized with the current robot pose, the remote server predicts an eight-step action chunk, and the robot executes the returned actions before the next image is acquired. The next inference request is therefore issued after completion of the current chunk in the physical experiments reported here. Parallel host-side inference was tested separately on the computer but was not used for the reported physical robot trials.

\subsection{Cloud-Edge Execution and Safety Filtering}

The 7B OpenVLA-OFT model runs on a remote inference server~\cite{Kim2024OpenVLA}, while the robotic client remains CPU-only. On the RTX 6000 Ada server used in our experiments, the logged model-side inference time was approximately 0.06--0.08~s per request. This measurement excludes image capture, network transmission, and physical robot execution time. At each replan step, the client sends $(I_t,l,H)$ through FastAPI, receives $C_t$, and decodes it into candidate actions. A candidate action is rejected if any of the following checks fail:
\begin{equation}
\begin{aligned}
    p_t+\Delta p_t\in\mathcal{B},\quad z_t+\Delta z_t\geq z_{\min}, \\
    \|\Delta p_t\|_{\infty}\leq d_{\max},\quad
    \|\Delta r_t\|_{\infty}\leq r_{\max}.
\end{aligned}
\end{equation}
Here $\mathcal{B}$ is the allowed workspace, $z_{\min}$ is the minimum build-plate clearance, $d_{\max}=5$~mm, and $r_{\max}=0.05$~rad for the current post-print retrieval configuration. The gripper command must be a valid binary state. The client also rejects NaN actions, network-timeout responses, stale chunks, and commands arriving after an emergency-stop flag. If a command is rejected, the remaining chunk is discarded and the client either requests a new chunk or halts the episode, depending on the experiment mode.


\section{Experiments}
\label{sec:experiments}

\subsection{Experimental Scope and Connection to the Method}

The experiments evaluate both trajectory-level prediction and physical closed-loop execution of the proposed ChunkVLA-AM pipeline. Following the formulation in Section~\ref{sec:preliminaries}, the policy receives a monocular image $I_t$ and a language instruction $l$, then predicts a short-horizon action chunk $A_t=[a_t,\ldots,a_{t+H-1}]$. The logged-trajectory analysis evaluates whether these chunks preserve the structure of expert AM handling trajectories after data conversion, action normalization, LoRA adaptation, and chunk construction, while a separate 42-trial physical evaluation measures end-to-end A-to-B transfer success. The trajectory plots therefore describe open-loop prediction consistency, whereas the robot trials provide a direct closed-loop deployment check under the same FR3 workcell configuration.

\subsection{Post-Print Retrieval Task}

The evaluation includes two color-conditioned manipulation tasks:
``catch the blue block'' and ``catch the red block.'' In each task,
the FR3 approaches the specified block at point A, closes the gripper,
transfers the target to point B, and releases it. The two tasks use
the same manipulation sequence but require the policy to associate
the instruction with the corresponding visual target. These controlled
A-to-B block-transfer tasks serve as reproducible proxies for post-print
part retrieval and transfer while evaluating language-conditioned
Cartesian trajectory prediction and physical execution.

\subsection{Evaluation Metrics}

For each logged frame, the policy receives the monocular observation and instruction and predicts an $H$-step action chunk. The predicted future positions are compared with the expert end-effector trajectory along the $x$, $y$, and $z$ axes. Because these plots are generated from logged states and not from closed-loop rollouts, the metrics describe open-loop trajectory consistency, not final task success.

The primary trajectory metric is axis-wise mean absolute error (MAE),
\begin{equation}
    \mathrm{MAE}_k=\frac{1}{N}\sum_{t=1}^{N}|p^{\mathrm{pred}}_{t,k}-p^{\mathrm{expert}}_{t,k}|.
\end{equation}

\subsection{Comparison of Adaptation Methods and Prediction Horizons}

To evaluate embodiment-specific adaptation and compare the evaluated policy configurations, we tested four models on the same FAIRINO FR3, camera setup, workspace, and test data. The standard single-step {OpenVLA (Zero-Shot)} and the 8-step action-chunked {OpenVLA-OFT (Zero-Shot)} received no FR3-specific training. The single-step adapted model, referred to as {Bridging the Pretrain-to-Real Gap}~\cite{bridging_gap}, and {ChunkVLA-AM (Ours)} were both adapted using the same FR3 demonstration dataset. ChunkVLA-AM uses 8-step action chunks. Because adaptation and prediction horizon differ across configurations, this comparison evaluates the complete configurations without isolating chunk length as a single controlled variable.

\begin{figure}[h]
  \centering
  \includegraphics[width=\linewidth]{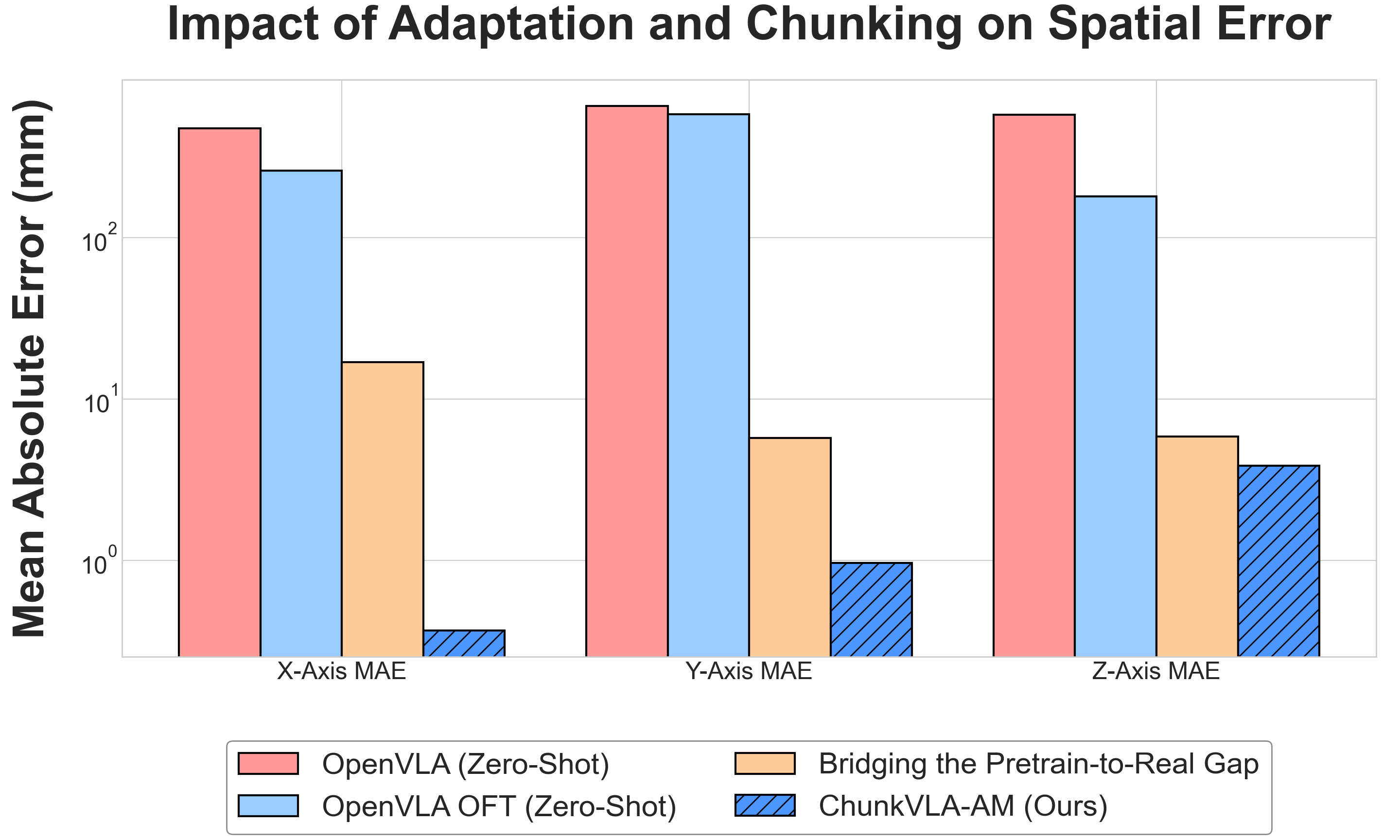}
  \caption{Trajectory tracking comparison across adaptation methods. The y-axis uses a logarithmic scale to highlight the significant precision gains. ChunkVLA-AM achieves sub-millimeter lateral accuracy, outperforming both zero-shot models and single-step fine-tuned baselines.}
  \label{fig:adaptation_comparison}
\end{figure}

We use Mean Absolute Error (MAE) across the X, Y, and Z axes as the primary evaluation metric to assess spatial tracking accuracy against the ground-truth Cartesian trajectory. The performance comparison is illustrated in Figure~\ref{fig:adaptation_comparison}, where a logarithmic scale is applied to the y-axis to accommodate the multi-order magnitude disparities.

The empirical data show that both zero-shot models are unsuitable for the FR3 additive manufacturing environment. OpenVLA (Zero-Shot) and OpenVLA-OFT (Zero-Shot) produce extreme spatial deviations, with MAE values ranging from 179.60~mm to 650.12~mm. This magnitude of error indicates that pre-trained visual backbones cannot directly transfer to the constrained workspace and specific kinematics of the FR3 robot without targeted training. The failure of the OpenVLA-OFT (Zero-Shot) model confirms that temporal action chunking alone cannot compensate for the lack of spatial semantics.

The single-step adaptation model~\cite{bridging_gap} significantly reduces the tracking error. However, it still yields an average spatial error of 9.50~mm across the three axes, with a pronounced deviation of 16.90~mm on the X-axis. In physical AM workcells, a 16.90~mm lateral offset easily causes collisions with the build plate or printed parts. This localized drift exposes the limitation of single-step autoregressive decoding, where slight execution deviations accumulate over time and lead to terminal open-loop drift.

In contrast, ChunkVLA-AM achieves an overall average MAE of 1.74~mm, with lateral errors of 0.37~mm on the X-axis and 0.97~mm on the Y-axis and a Z-axis error of 3.87~mm. The adapted chunked configuration therefore exhibits substantially lower trajectory error than the evaluated single-step adapted baseline. Because the compared configurations differ in both adaptation and prediction horizon, these results support the combined FR3-adaptation and chunked-deployment configuration but do not by themselves attribute the improvement solely to action chunking.

\subsection{Physical Closed-Loop Robot Validation}

We additionally evaluated the complete camera--server--robot loop in 42 physical A-to-B transfer trials using eight-step action chunks. The trials were evenly divided between the red and blue targets, with 21 trials per color. Across all 42 trials, 39 transfers were completed successfully, corresponding to an overall success rate of 92.9\%. The three failures occurred during terminal placement at point B: the placement height was not sufficiently controlled, causing the released object to fall and topple. These failures were retained in the reported success rate. The physical trials use the same FR3 workcell and sequential observation--inference--execution loop described in Section~\ref{sec:method}.

\subsection{Influence of Physical Illumination and Color Temperature}

\begin{figure*}[!t]
\centering
\includegraphics[width=0.84\textwidth]{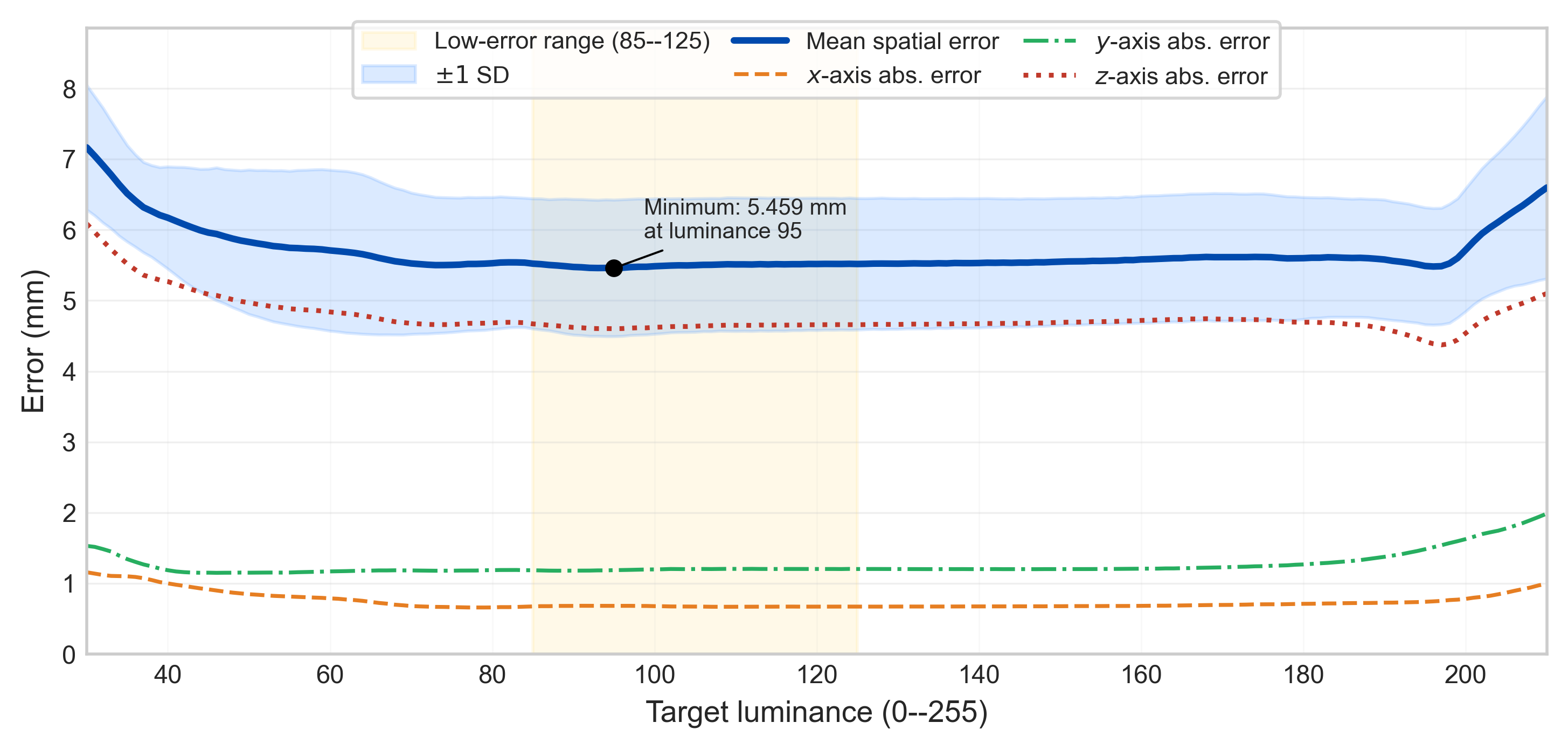}
\caption{Trajectory prediction errors across ten luminance-sweep trials. The solid curve and surrounding band show the across-trial mean spatial error and $\pm1$ standard deviation, respectively. The remaining curves show the axis-wise MAEs. The highlighted range from 85 to 125 indicates a low-error region centered around the minimum error observed at luminance 95.}
\label{fig:luminance_sweep}
\end{figure*}

We also evaluated ten separately recorded but task-matched logged test trajectories at 181 target luminance levels from 30 to 210; the ten trials are similar A-to-B sequences collected independently, not repeated copies of a single trajectory. As shown in Fig.~\ref{fig:luminance_sweep}, the mean spatial error reached its minimum of 5.459~mm at luminance 95 and remained low within the highlighted 85--125 range. The error increased to 7.165~mm and 6.598~mm at luminance 30 and 210, respectively. The axis-wise curves further show that the $z$-axis was the dominant source of error, whereas the $x$- and $y$-axis errors remained smaller.

\begin{figure}[!t]
\centering
\begin{minipage}{0.31\linewidth}
  \includegraphics[width=\linewidth]{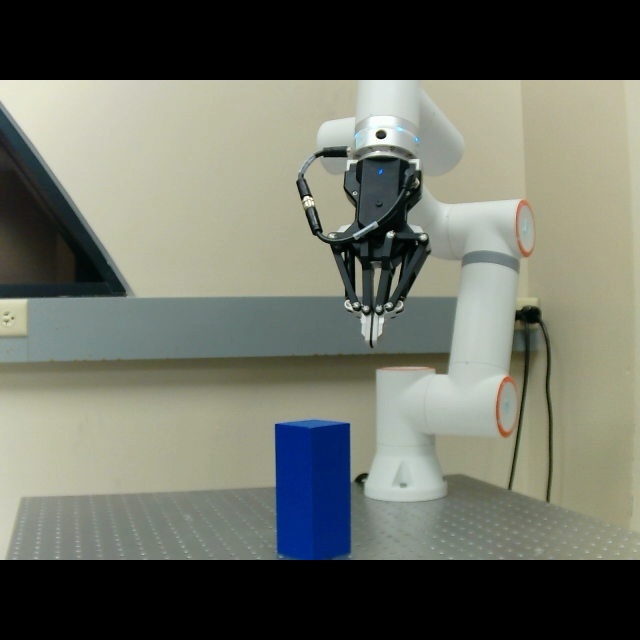}
  \centerline{\scriptsize (a) Baseline}
\end{minipage}\hfill
\begin{minipage}{0.31\linewidth}
  \includegraphics[width=\linewidth]{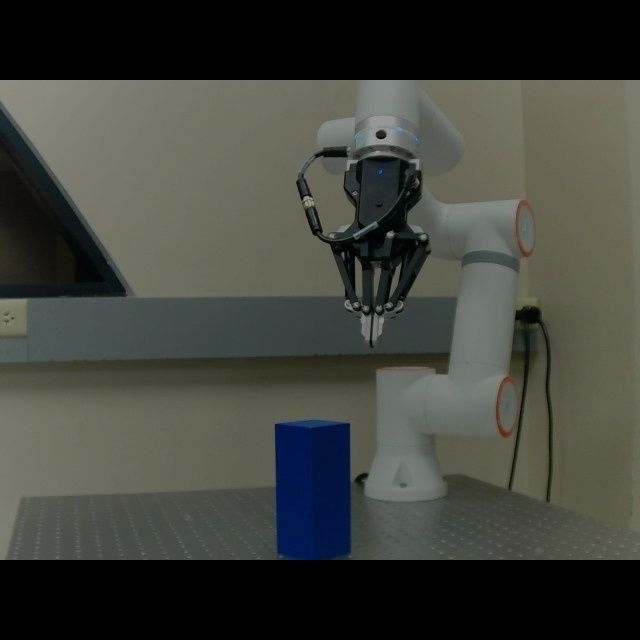}
  \centerline{\scriptsize (b) LowLight}
\end{minipage}\hfill
\begin{minipage}{0.31\linewidth}
  \includegraphics[width=\linewidth]{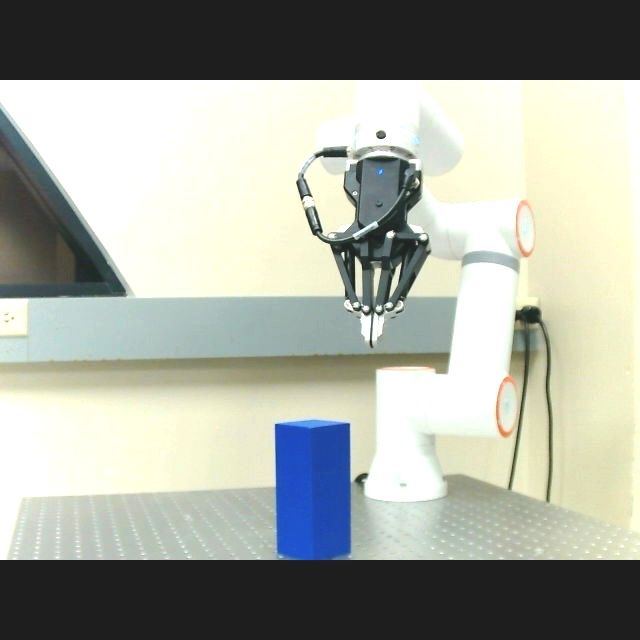}
  \centerline{\scriptsize (c) OverLight Mild}
\end{minipage}

\vspace{2pt}

\begin{minipage}{0.31\linewidth}
  \includegraphics[width=\linewidth]{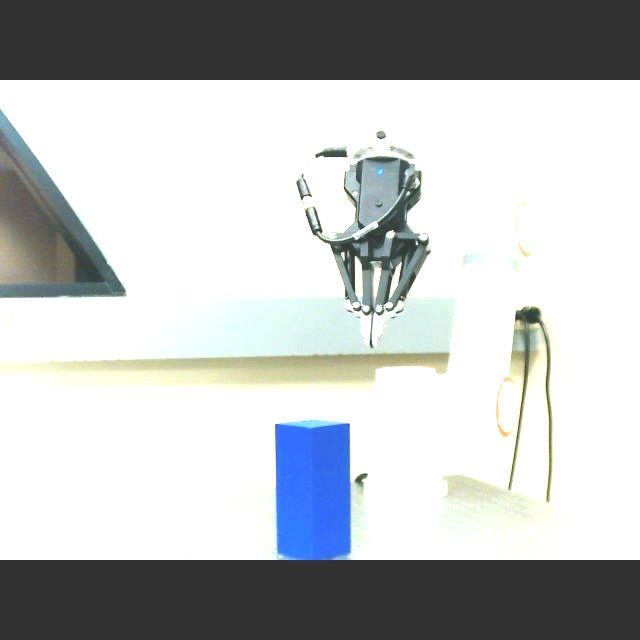}
  \centerline{\scriptsize (d) OverLight Extreme}
\end{minipage}\hfill
\begin{minipage}{0.31\linewidth}
  \includegraphics[width=\linewidth]{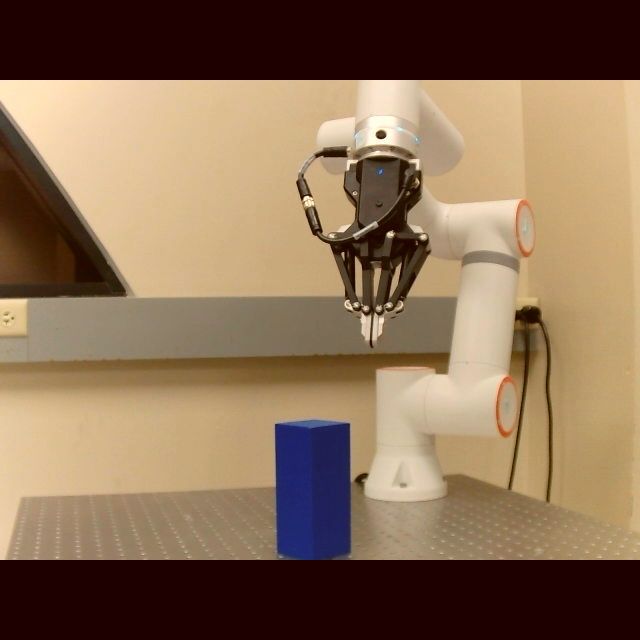}
  \centerline{\scriptsize (e) ColorTemp Warm}
\end{minipage}\hfill
\begin{minipage}{0.31\linewidth}
  \includegraphics[width=\linewidth]{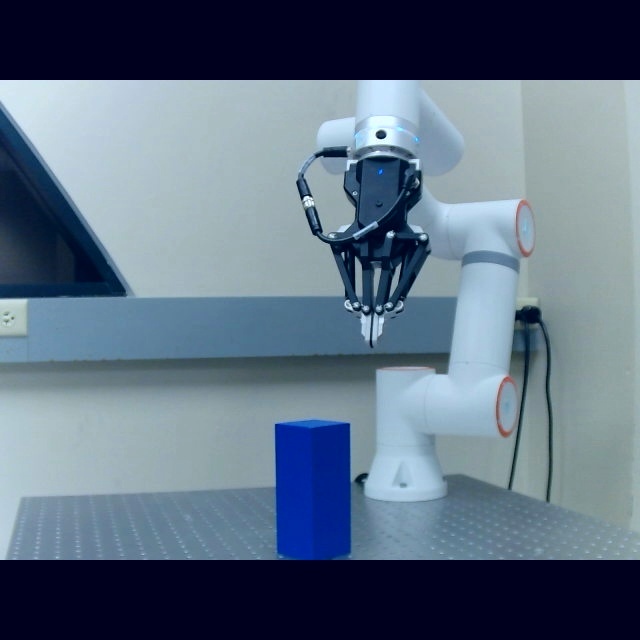}
  \centerline{\scriptsize (f) ColorTemp Cold}
\end{minipage}
\caption{Representative FR3 workspace captures under six illumination conditions used in the lighting robustness study. Panel (a) is the neutral baseline; the remaining panels span low light, over-exposure, and warm/cold color-temperature variants. Luminance values are mean pixel intensity on a 0--255 scale.}
\label{fig:lighting_samples}
\end{figure}

The experimental setup used the same FR3 workspace under controlled physical lighting changes. Illumination was varied by adjusting the room light intensity and swapping between neutral white, warm (3000\,K), and cold (6500\,K) light sources. All captures were taken with a fixed monocular RGB camera at the same pose; no digital post-processing or gamma correction was applied. Figure~\ref{fig:lighting_samples} shows six representative captures from the seven tested illumination conditions, spanning variations in luminance and color temperature.

The policy was evaluated directly on environmental captures under Low Light, Baseline, and Over-Exposed conditions without mathematically simulating brightness. As shown in Table~\ref{tab:brightness_comparison}, the system maintains similar spatial tracking error across mild lighting shifts (luminance spanning 65.2 to 165.1), with the $z$-axis MAE remaining approximately 3.8--4.0~mm.

However, extreme over-exposure (luminance 189.1) degrades lateral tracking, with the $y$-axis MAE increasing to 2.30~mm. The warm and cold color-temperature conditions remain close to the baseline, while extreme glare produces the largest degradation. This result motivates additional lighting augmentation for highly reflective AM workspaces.

\begin{table}[h]
\centering
\caption{Influence of environmental illumination and color temperature on chunked trajectory prediction. Brightness is measured in mean luminance on a 0--255 scale. Errors are reported in axis-wise Mean Absolute Error (MAE) alongside the 3D Euclidean error ($L_2$).}
\label{tab:brightness_comparison}
\begin{tabular}{l|c|cccc}
\hline
\textbf{Condition} & \textbf{Luminance} & \multicolumn{4}{c}{\textbf{MAE (mm)}} \\
& \textbf{(0--255)} & $x$ & $y$ & $z$ & \textbf{3D $L_2$} \\
\hline
Baseline (Neutral) & 109 & 0.37 & \textbf{0.97} & 3.87 & 4.01 \\
LowLight Extreme & 30 & 0.45 & 1.28 & 4.04 & 4.26 \\
LowLight Mild & 65 & \textbf{0.26} & 1.03 & 4.03 & 4.17 \\
OverLight Mild & 165 & 0.37 & 1.00 & 3.86 & 4.00 \\
OverLight Extreme & 190 & 0.59 & 2.30 & 4.65 & 5.22 \\
\hline
ColorTemp Warm & 116 & 0.41 & 0.99 & 3.93 & 4.07 \\
ColorTemp Cold & 108 & 0.50 & \textbf{0.97} & \textbf{3.76} & \textbf{3.92} \\
\hline
\end{tabular}
\end{table}

We further analyzed a dense luminance sweep comprising 10 trials, each
evaluated at 181 target luminance levels from 30 to 210. The realized
image luminance closely followed the requested value: the mean
actual-minus-target difference was 0.0037 and the maximum absolute
difference was 0.0742 on the 0--255 scale. Because the sweep outputs
contain neither a color-temperature field nor a task-success label,
this analysis is used specifically to quantify trajectory-error
sensitivity to luminance and does not replace the warm/cold comparison
in Table~\ref{tab:brightness_comparison}.

Figure~\ref{fig:luminance_sweep} visualizes the low-error region around
the aggregate minimum and the error increases near the darkest and
brightest ends of the sweep.

\begin{table}[!t]
\centering
\caption{Trial-level summary of the dense luminance sweep. The reported
sweep error is the millimeter-valued spatial-error metric provided in
the analysis outputs and is kept distinct from the 3D $L_2$ values in
Table~\ref{tab:brightness_comparison}. Region means are first computed
within each of the 10 trials.}
\label{tab:luminance_sweep_summary}
\setlength{\tabcolsep}{3.5pt}
\begin{tabular}{l|c|ccc}
\hline
\textbf{Luminance region} & \textbf{Sweep error} &
\multicolumn{3}{c}{\textbf{Axis MAE (mm)}} \\
\textbf{(0--255)} & \textbf{(mm)} & $x$ & $y$ & $z$ \\
\hline
30--84   & 5.872 & 0.834 & 1.209 & 4.985 \\
85--125  & \textbf{5.499} & \textbf{0.676} &
           \textbf{1.198} & \textbf{4.639} \\
126--210 & 5.644 & 0.720 & 1.331 & 4.688 \\
\hline
\end{tabular}
\end{table}

The dense sweep supports the 85--125 interval as a narrower low-error
band without implying a single optimal illumination. Its
trial-level mean spatial error was 5.499~mm, whereas the 30--84 and
126--210 regions increased the error by 0.373~mm (6.8\%) and
0.146~mm (2.6\%), respectively; descriptive bootstrap 95\% confidence
intervals for these paired regional differences were [0.225, 0.529]~mm
and [0.064, 0.233]~mm. Within 85--125, the aggregate mean curve varied
by only 1.2\% of the global minimum. Although the across-trial curve
reached its lowest value of 5.459~mm at luminance 95 and remained
within 5\% of this minimum from 58 to 200, the error-minimizing
luminance of individual trials ranged from 36 to 198 (median 95).
Thus, luminance 95 should not be interpreted as a universal optimum.

Degradation became clearer at the sweep endpoints: the aggregate mean
error was 7.165~mm at luminance 30 and 6.598~mm at luminance 210,
corresponding to increases of 31.2\% and 20.9\% over the aggregate
minimum. The axis-wise results in
Table~\ref{tab:luminance_sweep_summary} show that the $z$-axis
remained the dominant error component throughout the sweep. Relative
to the 85--125 band, low-light degradation was primarily associated
with larger $z$- and $x$-axis errors, whereas the largest
high-luminance increase occurred on the $y$-axis. This continuous-sweep
evidence is consistent with the discrete-condition results:
intermediate illumination produces lower error, while the darkest and
brightest regimes are more likely to disturb visual grounding.

Overall, illumination changes affect height and lateral localization differently in the fixed AM workcell, highlighting the importance of lighting robustness during deployment.

\section{Conclusion and Discussion}
\label{sec:conclusion}

This paper presented ChunkVLA-AM as a reproducible OpenVLA-OFT-based system pipeline for FR3-based post-print retrieval in additive manufacturing. The paper specifies the data conversion interface, 7-D Cartesian action representation, horizon-based chunk construction, LoRA adaptation setup, cloud-edge request loop, and robot-side action filtering needed to reproduce the system. The logged-trajectory analyses show low Cartesian prediction error for the FR3-adapted eight-step configuration, while the physical evaluation completed 39 of 42 A-to-B transfer trials successfully (92.9\%). The three failures occurred during terminal placement when the released object toppled because of insufficient placement-height control. The lighting experiments further show that prediction error remains lowest over an intermediate luminance range and increases toward the darkest and brightest conditions. These results support the feasibility of the proposed deployment pipeline in the evaluated fixed FR3 workcell. Future work will add broader object and task generalization, controlled chunk-length ablations, contact-force monitoring, and richer AM-specific tasks such as warped-part inspection and failed-print removal.

\section*{Acknowledgment}

The authors gratefully acknowledge funding from the National Science Foundation (NSF) CREST Center for Multidisciplinary Research Excellence in Cyber-Physical Infrastructure Systems (MECIS) under Award No. 2112650. Additional support was provided by the NSF Expand AI PARTNER: ARISE: AI Research and Innovation for Smart Environments under NSF Award No. 2434916, as well as NSF ACCESS: AI-Enhanced Cross-Scale Sensing and Intelligent Quality Assessment for Scalable Additive Manufacturing (NSF Award No. ELE250047). The authors also acknowledge the University Transportation Center for Railway Safety (UTCRS) for providing the datasets used in this work and funding support under the USDOT UTC Program Grant No. 69A3552348340.

\bibliographystyle{IEEEtran}
\bibliography{example}

@inproceedings{Brohan2022RT1,
  title        = {{RT}-1: Robotics Transformer for Real-World Control at Scale},
  author       = {Brohan, Anthony and others},
  booktitle    = {Proceedings of Robotics: Science and Systems},
  address      = {Daegu, Republic of Korea},
  month        = jul,
  year         = {2023},
  doi          = {10.15607/RSS.2023.XIX.025},
  url          = {https://roboticsproceedings.org/rss19/p025.html}
}

@inproceedings{Brohan2023RT2,
  title        = {{RT}-2: Vision-Language-Action Models Transfer Web Knowledge to Robotic Control},
  author       = {Zitkovich, Brianna and others},
  booktitle    = {Proceedings of the 7th Conference on Robot Learning},
  series       = {Proceedings of Machine Learning Research},
  volume       = {229},
  pages        = {2165--2183},
  year         = {2023},
  publisher    = {PMLR},
  url          = {https://proceedings.mlr.press/v229/zitkovich23a.html}
}

@inproceedings{Kim2024OpenVLA,
  title        = {{OpenVLA}: An Open-Source Vision-Language-Action Model},
  author       = {Kim, Moo Jin and others},
  booktitle    = {Proceedings of the 8th Conference on Robot Learning},
  series       = {Proceedings of Machine Learning Research},
  volume       = {270},
  pages        = {2679--2713},
  year         = {2025},
  publisher    = {PMLR},
  url          = {https://proceedings.mlr.press/v270/kim25c.html}
}

@inproceedings{OpenXEmbodiment2023,
  title        = {{Open X-Embodiment}: Robotic Learning Datasets and {RT}-{X} Models},
  author       = {{Open X-Embodiment Collaboration} and others},
  booktitle    = {IEEE International Conference on Robotics and Automation ({ICRA})},
  pages        = {6892--6903},
  year         = {2024},
  doi          = {10.1109/ICRA57147.2024.10611477}
}

@article{Reed2022Gato,
  title        = {A Generalist Agent},
  author       = {Reed, Scott and others},
  journal      = {Transactions on Machine Learning Research},
  year         = {2022},
  url          = {https://openreview.net/forum?id=1ikK0kHjvj}
}

@inproceedings{Ahn2022SayCan,
  title        = {Do As {I} Can, Not As {I} Say: Grounding Language in Robotic Affordances},
  author       = {Ichter, Brian and others},
  booktitle    = {Proceedings of the 6th Conference on Robot Learning},
  series       = {Proceedings of Machine Learning Research},
  volume       = {205},
  pages        = {287--318},
  year         = {2023},
  publisher    = {PMLR},
  url          = {https://proceedings.mlr.press/v205/ichter23a.html}
}

@inproceedings{Driess2023PaLME,
  title        = {{PaLM-E}: An Embodied Multimodal Language Model},
  author       = {Driess, Danny and others},
  booktitle    = {Proceedings of the 40th International Conference on Machine Learning},
  series       = {Proceedings of Machine Learning Research},
  volume       = {202},
  pages        = {8469--8488},
  year         = {2023},
  publisher    = {PMLR},
  url          = {https://proceedings.mlr.press/v202/driess23a.html}
}

@inproceedings{Jiang2022VIMA,
  title        = {{VIMA}: Robot Manipulation with Multimodal Prompts},
  author       = {Jiang, Yunfan and others},
  booktitle    = {Proceedings of the 40th International Conference on Machine Learning},
  series       = {Proceedings of Machine Learning Research},
  volume       = {202},
  pages        = {14975--15022},
  year         = {2023},
  publisher    = {PMLR},
  url          = {https://proceedings.mlr.press/v202/jiang23b.html}
}

@article{Guadarrama2021RLDS,
  title        = {{RLDS}: An Ecosystem to Generate, Share and Use Datasets in Reinforcement Learning},
  author       = {Ramos, Sabela and others},
  journal      = {arXiv preprint arXiv:2111.02767},
  year         = {2021},
  doi          = {10.48550/arXiv.2111.02767},
  url          = {https://arxiv.org/abs/2111.02767}
}

@misc{TFDS,
  title        = {{TensorFlow Datasets}: A Collection of Ready-to-use Datasets},
  url          = {https://www.tensorflow.org/datasets},
  note         = {Accessed: Jun. 23, 2026}
}

@inproceedings{Hu2022LoRA,
  title        = {{LoRA}: Low-Rank Adaptation of Large Language Models},
  author       = {Hu, Edward J. and others},
  booktitle    = {International Conference on Learning Representations},
  year         = {2022},
  url          = {https://openreview.net/forum?id=nZeVKeeFYf9}
}

@book{Gibson2021AM,
  title        = {Additive Manufacturing Technologies},
  author       = {Gibson, Ian and Rosen, David and Stucker, Brent and Khorasani, Mahyar},
  publisher    = {Springer},
  address      = {Cham, Switzerland},
  edition      = {3},
  year         = {2021},
  doi          = {10.1007/978-3-030-56127-7}
}

@article{Ngo2018AMReview,
  title        = {Additive Manufacturing ({3D} Printing): A Review of Materials, Methods, Applications and Challenges},
  author       = {Ngo, Tuan D. and Kashani, Alireza and Imbalzano, Gabriele and Nguyen, Kate T. Q. and Hui, David},
  journal      = {Composites Part B: Engineering},
  volume       = {143},
  pages        = {172--196},
  year         = {2018},
  doi          = {10.1016/j.compositesb.2018.02.012}
}

@article{DebRoy2018MetalAM,
  title        = {Additive Manufacturing of Metallic Components -- Process, Structure and Properties},
  author       = {DebRoy, Tarasankar and others},
  journal      = {Progress in Materials Science},
  volume       = {92},
  pages        = {112--224},
  year         = {2018},
  doi          = {10.1016/j.pmatsci.2017.10.001}
}

@article{Ding2015WireArcAM,
  title        = {Wire-Feed Additive Manufacturing of Metal Components: Technologies, Developments and Future Interests},
  author       = {Ding, Donghong and Pan, Zengxi and Cuiuri, Dominic and Li, Huijun},
  journal      = {The International Journal of Advanced Manufacturing Technology},
  volume       = {81},
  pages        = {465--481},
  year         = {2015},
  doi          = {10.1007/s00170-015-7077-3}
}

@article{Stephens2013UFP,
  title        = {Ultrafine Particle Emissions from Desktop {3D} Printers},
  author       = {Stephens, Brent and Azimi, Parham and El Orch, Zeineb and Ramos, Tiffanie},
  journal      = {Atmospheric Environment},
  volume       = {79},
  pages        = {334--339},
  year         = {2013},
  doi          = {10.1016/j.atmosenv.2013.06.050}
}

@article{Azimi2016FDMEmissions,
  title        = {Emissions of Ultrafine Particles and Volatile Organic Compounds from Commercially Available Desktop Three-Dimensional Printers with Multiple Filaments},
  author       = {Azimi, Parham and Zhao, Dan and Pouzet, Christelle and Crain, Neil E. and Stephens, Brent},
  journal      = {Environmental Science \& Technology},
  volume       = {50},
  number       = {3},
  pages        = {1260--1268},
  year         = {2016},
  doi          = {10.1021/acs.est.5b04983}
}

@article{Chen2020MetalAMHazards,
  title        = {Exposure, Assessment and Health Hazards of Particulate Matter in Metal Additive Manufacturing: A Review},
  author       = {Chen, Rui and others},
  journal      = {Chemosphere},
  volume       = {259},
  pages        = {127452},
  year         = {2020},
  doi          = {10.1016/j.chemosphere.2020.127452}
}

@article{Zhang2006FDMFEM,
  title        = {Three-Dimensional Finite Element Analysis Simulations of the Fused Deposition Modelling Process},
  author       = {Zhang, Y. and Chou, Y. K.},
  journal      = {Proceedings of the Institution of Mechanical Engineers, Part B: Journal of Engineering Manufacture},
  volume       = {220},
  number       = {10},
  pages        = {1663--1671},
  year         = {2006},
  doi          = {10.1243/09544054JEM572}
}

@article{Jin2020InterlayerAI,
  title        = {Automated Real-Time Detection and Prediction of Interlayer Imperfections in Additive Manufacturing Processes Using Artificial Intelligence},
  author       = {Jin, Zeqing and Zhang, Zhizhou and Gu, Grace X.},
  journal      = {Advanced Intelligent Systems},
  volume       = {2},
  number       = {1},
  pages        = {1900130},
  year         = {2020},
  doi          = {10.1002/aisy.201900130}
}

@article{Zhang2017SLMDefects,
  title        = {Defect Formation Mechanisms in Selective Laser Melting: A Review},
  author       = {Zhang, Bi and Li, Yongtao and Bai, Qian},
  journal      = {Chinese Journal of Mechanical Engineering},
  volume       = {30},
  pages        = {515--527},
  year         = {2017},
  doi          = {10.1007/s10033-017-0121-5}
}

@inproceedings{Zhao2023ACT,
  title        = {Learning Fine-Grained Bimanual Manipulation with Low-Cost Hardware},
  author       = {Zhao, Tony Z. and Kumar, Vikash and Levine, Sergey and Finn, Chelsea},
  booktitle    = {Proceedings of Robotics: Science and Systems},
  address      = {Daegu, Republic of Korea},
  month        = jul,
  year         = {2023},
  doi          = {10.15607/RSS.2023.XIX.016},
  url          = {https://roboticsproceedings.org/rss19/p016.html}
}

@inproceedings{Chi2023DiffusionPolicy,
  title        = {Diffusion Policy: Visuomotor Policy Learning via Action Diffusion},
  author       = {Chi, Cheng and others},
  booktitle    = {Proceedings of Robotics: Science and Systems},
  address      = {Daegu, Republic of Korea},
  month        = jul,
  year         = {2023},
  doi          = {10.15607/RSS.2023.XIX.026},
  url          = {https://roboticsproceedings.org/rss19/p026.html}
}

@inproceedings{bridging_gap,
  title        = {Bridging the Pretrain-to-Real Gap: Alignment Challenges in Deploying Generalist {VLA} Models for Additive Manufacturing},
  author       = {Liu, Zhugang and others},
  booktitle    = {Proceedings of the IEEE/CVF Conference on Computer Vision and Pattern Recognition ({CVPR}) Workshops},
  pages        = {1048--1056},
  month        = jun,
  year         = {2026}
}

\end{document}